%% file: main.tex
\documentclass{article}
\usepackage{ijcai22}

\usepackage{times}

\usepackage{soul}
\usepackage{url}
\usepackage[hidelinks]{hyperref}
\usepackage[utf8]{inputenc}
\usepackage[small]{caption}
\usepackage{graphicx}
\usepackage{amsmath}
\usepackage{booktabs}
\usepackage{algorithm}
\usepackage{algorithmic}
\usepackage{mathrsfs}
\usepackage{multirow}
\usepackage{xspace}
\usepackage{subfigure} 
\usepackage{hyperref}
\usepackage{amsmath}
\usepackage{amssymb}
\usepackage{amsfonts}
\usepackage{xcolor}
\usepackage{epsfig}
\usepackage{diagbox}

\title{Feature and Instance Joint Selection: A Reinforcement Learning Perspective}

\author{
Wei Fan$^1$\and
Kunpeng Liu$^1$\and
Hao Liu$^{2}$\and
Hengshu Zhu$^3$\and
Hui Xiong$^{4\,*}$\And
Yanjie Fu$^1$\footnote{Corresponding Author}\\
\affiliations
$^1$University of Central Florida, $^2$Hong Kong University of Science and Technology, \\ 
$^3$Baidu Talent Intelligence Center, $^4$Rutgers University \\
\emails
\{weifan, kunpengliu\}@knights.ucf.edu,
liuh@ust.hk,
zhuhengshu@gmail.com,\\
hxiong@rutgers.edu,
yanjie.fu@ucf.edu
}

\begin{document}

\maketitle

\begin{abstract}

\input{abstract}
\end{abstract}

\input{intro}
\input{preliminary}

\input{method}

\input{experiment}

\input{related_work}

\input{conclusion}

\bibliographystyle{named}
\bibliography{ijcai22.bib}

\end{document}

%% file: abstract.tex
Feature selection and instance selection are two important techniques of data processing. However, such selections have mostly been studied separately, while existing work towards the joint selection conducts feature/instance selection coarsely; thus neglecting the latent fine-grained interaction between feature space and instance space. To address this challenge, we 
propose a reinforcement learning solution to accomplish the joint selection task and simultaneously capture the interaction between the selection of each feature and each instance.
In particular, a sequential-scanning mechanism is designed as action strategy of agents and a collaborative-changing environment is used to enhance agent collaboration.
In addition, an interactive paradigm introduces prior selection knowledge to help agents for more efficient exploration. Finally, extensive experiments on real-world datasets have demonstrated the improved performances. 

%% file: intro.tex
\section{Introduction}


Data preprocessing is to make input data most appropriate for model training.
Generally, two well-known preprocessing techniques are feature selection \cite{liu2012feature} and instance selection \cite{brighton2002advances}.
Feature selection is to select most important features to increase predictive performance (e.g., accuracy) and reduce feature size.
Instance selection shares the similar objective to simultaneously improve modeling accuracy and decrease instance size.


In prior literature,  feature selection and instance selection
are usually regarded as two separate problems. Limited algorithms have been proposed for their joint selection.
The joint feature and instance selection was initially addressed using genetic algorithms \cite{kuncheva1999nearest}. Then some studies solve this problem with principal component analysis \cite{suganthi2019instance} or application of pairwise similarity \cite{benabdeslem2020scos}. Others try to apply heuristic search and adopt simulated annealing algorithms \cite{de2008novel} or sequential forward search \cite{garcia2021si}.
In fact, the selection of each feature and each instance are mutually influenced and jointly decide the final selection results.
However, most previous work of joint selection conducts feature/instance selection coarsely and neglects the fine-grained interaction of feature space and instance space, which largely hinders the predictive performances on the selected data.



Reinforcement Learning (RL), as an effective tool, has great potential to learn the optimal results for search problems like such selections \cite{liu2019automating}.
In particular, different agents 
(i) individually take actions step by step that is appropriate to model each fine-grained selection choice; 
(ii) mutually interact with each other that can capture the interaction between feature space and instance space;
(iii) jointly targets for optimal decisions that can be regarded as the selected data of joint selection task.
To this end, we make a first attempt to leverage reinforcement learning for the joint feature instance selection problem. For this goal, several challenges arise.

First, how to formulate the joint feature instance selection task with reinforcement learning?
Feature/Instance selection usually repeats two steps: select a subset and test the performance. If viewed from the RL perspective, this exploratory process is an agent first selects features/instances and then observe the selected data to get reward. 
Considering the two selections, we naturally reformulate the joint selection with a dual-agent reinforcement learning paradigm. 
Specifically, we create two RL agents: 
1) a feature agent aims to select the optimal feature subset; 2) an instance agent aims to select the optimal instance subset.
The two agents perceive selected features and instances as the state of environment, collect data characteristics as reward, and interact with each other to search for optimum selection results.

Second, how can we enable the two agents to simultaneously and collaboratively conduct joint selection? 
On one hand, if the feature or instance agent selects a subset each time, the agent needs to make $n$ binary selections on $n$ features/instances. This results into action space is exponentially-increasing ($2^n$) with the number of features/instances, leading to practical implementation difficulties. 
Thus, we propose a {\textit{sequential-scanning}} mechanism for action design of agents. Specifically, we organize the selection decisions of features as a sequence and let the feature agent iteratively scan over this sequence to (de)select one feature each time. The instance agent adopts the same scanning strategy. 
This mechanism significantly reduces action space from exponential ($2^n$) to binary choice and transforms selection as a sequential decision-making process, which is appropriate for RL.
On the other hand, the performance of a downstream model highly depends on the joint quality of both features and instances.
To achieve the global optimal of joint selection, the two agents need to collaborate to learn the mutual influence of features and instances. We thus develop a {\textit{collaborative-changing}} environment for agents.
We regard the environment as the selected data sub-matrix, where the columns (features) and rows (instances) are simultaneously changed by the actions of dual agents.
This shared environment captures actions of the two agents and jointly sense the data quality in two dimensions.



Third, how can the two agents learn prior knowledge to improve learning efficiency?
Interactive RL \cite{amir2016interactive} has shown its superiority on speeding up agent exploration by learning from human experts or prior knowledge.
In this regard, we utilize two external trainers to teach the two agents respectively via interactive RL:  we introduce
1) a random forest based trainer with knowledge of feature importance to teach feature agent to pick features. 
2) an isolation forest based trainer to recognize instance anomaly to teach instance agent how to filter out instances. 
With the advice of the two trainers, the two agents can learn the patterns of spotting and selecting quality features and instances more efficiently. 

In summary, we propose a dual-agent interactive reinforcement learning framework to model the interaction of the joint feature and instance selection. 
Our contributions are:
(i) we formulate the joint selection task with dual-agent reinforcement learning;
(ii) we propose a sequential-scanning mechanism and a collaborative-changing environment to achieve the simultaneous and interactive selection;   
(iii) we leverage interactive reinforcement learning to improve the learning efficiency;
(iv) we conduct extensive experiments to demonstrate our improved performances.

%% file: preliminary.tex
\section{Backgrounds}

\noindent\textit{Concept 2.1} \textbf{Dual-Agent Reinforcement Learning} is a variant of multi-agent reinforcement learning.  The dual-agent RL has two agents collaboratively accomplish two different tasks. 
For example, it has been applied to interactively generate bounding boxes and detect facial landmarks in computer vision~\cite{guo2018dual}.


\noindent\textit{Concept 2.2} \textbf{Interactive Reinforcement Learning} is to provide agents with action advice from teacher-like trainers, so that agents learn learn the optimal decisions more efficiently in early exploration~\cite{amir2016interactive}.

\noindent\textit{Definition 2.3} \textbf{Feature Selection}. Given an input data matrix $ {X}  \in {\mathbb R}^{n*m}$, $n$ and $m$ denote the number of instances and features respectively; $x_{ij}$ is the element in the $i$-th row and $j$-th column. We denote input features $ {F} = \{ \mathbf{f}_j \}_{j=1}^{m}$,  
where the $j$-th feature is denoted by ${\mathbf{f}_j} = \{x_{1j},x_{2j}, ..., x_{nj} \}$.
Feature selection aims to select an optimal subset ${F^*} \subseteq {F}$ making downstream predictive model perform well.

\noindent\textit{Definition 2.4} \textbf{Instance Selection}. As we mentioned above, for the data matrix ${X}$, all the instances are denoted as ${C} = \{ \mathbf{c}_i \}_{i=1}^{n}$
where the $i$-th instance $\mathbf{c}_i =\{x_{i1},x_{i2}, ..., x_{im} \}$. Traditionally, instance selection studies aim to remove data noise or outliers, and find an instance subset ${C^*} \subseteq {C}$ that has the same performances as ${C}$ for the downstream predictive task \cite{wilson2000reduction}.

\noindent\textit{Definition 2.5} \textbf{The Joint Feature and Instance Selection Task} is to simultaneously find the optimal feature subset ${F^*}$ and the optimal instance subset ${C^*}$, in order to achieve the best performances in a downstream predictive task.





%% file: method.tex
\section{Method}
We address aforementioned challenges and propose a framework which called Dual-Agent Interactive Reinforced Selection (DAIRS), to model joint feature and instance selection task and introduce prior selection knowledge to agents for interactive reinforcement learning.
\subsection{DAIRS}
As a reinforcement learning based framework, the DAIRS framework consists of dual agents, actions, states, reward and trainers. Specifically, 

\subsubsection{Dual Agents.}
The two agents are:
the \textit{ Feature Agent} which models feature-feature correlation to select an optimized feature subset; 
the \textit{ Instance Agent} which models instance-instance correlation to select an optimized instance subset. 
However, the local optimal in feature or instance selection can't guarantee the global optimal of joint feature-instance selection.
As a result, the two agents need to strategically collaborate and transfer knowledge between feature and instance selections. 

\subsubsection{Actions.}
\begin{figure}[h]
\centering
\includegraphics[width=8.35cm]{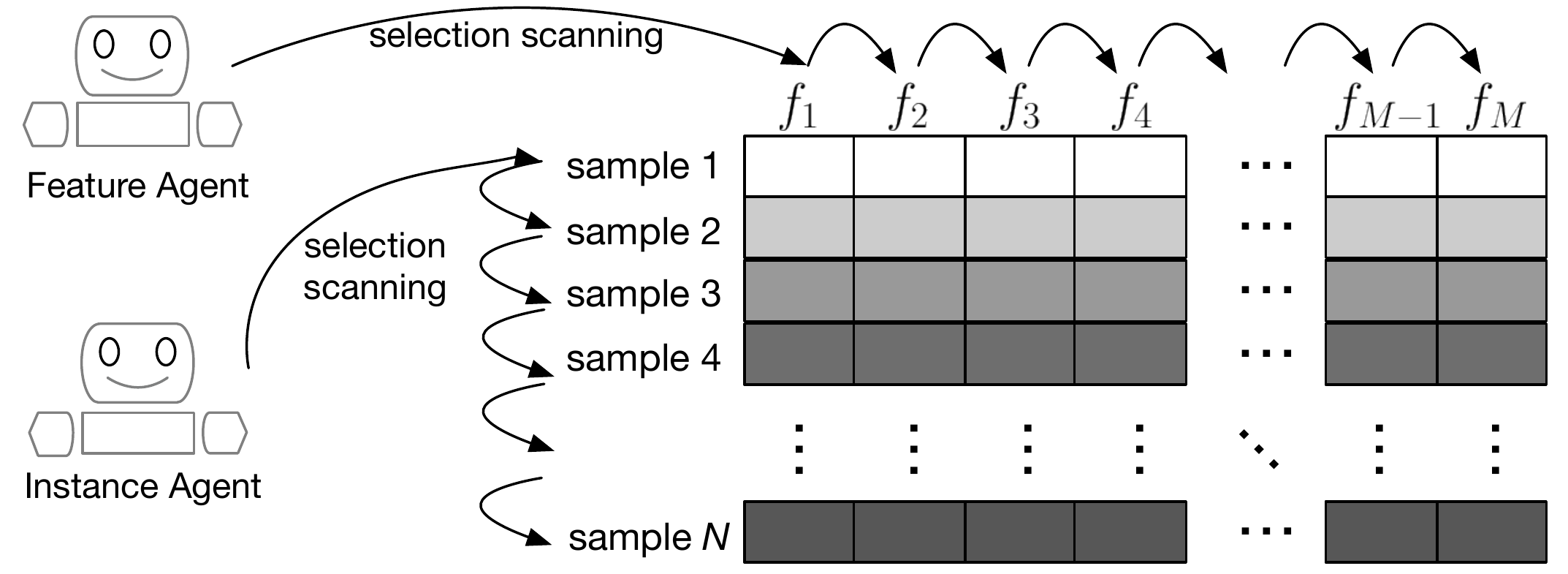}
\vspace{-0mm}
\caption{Actions of dual agents with sequential scanning. Agents iteratively scan over each feature or each instance to make the select/deselect decisions.}
\label{method1}
\vspace{-2mm}
\end{figure}

The dual agent action design is critical, because we need to consider: (i) the action space that determines the computational complexity and the learning efficiency of agents; (ii) the fine-grained interaction between feature space and instance space. 
The two considerations make it impossible to directly apply classic multi-agent reinforced selection~\cite{liu2019automating} to joint features and instance selection. 
To tackle this challenge, we develop a {\textit{sequential-scanning }} with restart mechanism  ({Figure \ref{method1}}). Specifically,

\noindent \underline{\textit{Actions of the feature agent}}: are to sequentially scan all the $m$ features and then restart scanning of the features over steps, where each step will select or deselect one feature. Let us denote a feature action record by $\mathbf{a_{F}} = \{a_i\}_{i=1}^{m}$, where the $i$-th action $a_i = 1$ or $a_i = 0$ means to select or deselect the $i$-th feature. Since the feature agent will restart the sequential scanning of $m$ features, at Step $t$, the feature agent decides to select or deselect $\mathbf{f}_{t\,(mod\, m)}$.

\noindent \underline{\textit{Actions of the instance agent}}: are to sequentially scan  all the $n$ instances, and then restart scanning of the instances over steps, where each step will select or deselect one instance. Let us denote an instance action record as $\mathbf{a_{I}} = \{a_i\}_{i=1}^{n}$, where the $i$-th action $a_i = 1$ or $a_i = 0$ means to select or deselect the instance $\mathbf{c}_i$. At Step $t$, the instance agent decides to select or deselect instance $\mathbf{c}_{t\,(mod\, n)}$.

The sequential scanning strategy allows features and instances to be simultaneously selected or deselected.
The dual interaction will coordinate the scanning actions  of the two agents to generate a globally optimized data subspace of instances and features. 

\subsubsection{State of the Environment.}
Instead of separately creating two environments for feature agent and instance agent, we  develop a sharing environment to support simultaneous  interaction between agents.
The state is to quantitatively represent the situation of the  {\textit{collaboratively-changing}} environment.
However, this is a non-trivial task: at the $t$-th step, the action records of the feature and instance agents, denoted by $\mathbf{a_F}^t$ and $ \mathbf{a_I}^t$, can respectively derive a feature subset ${F'}$  and an instance subset ${C'}$. 
The two subsets jointly form a sub-matrix of the input data ${X}$. 
The challenge is that the selected sub-matrix ${X'}$ cannot directly be regarded as the state, because its dimensions change dynamically over time, while learning the policy networks of the dual agents require a fixed-length state representation.

\begin{figure}[h]
\centering
\includegraphics[width=8.6cm]{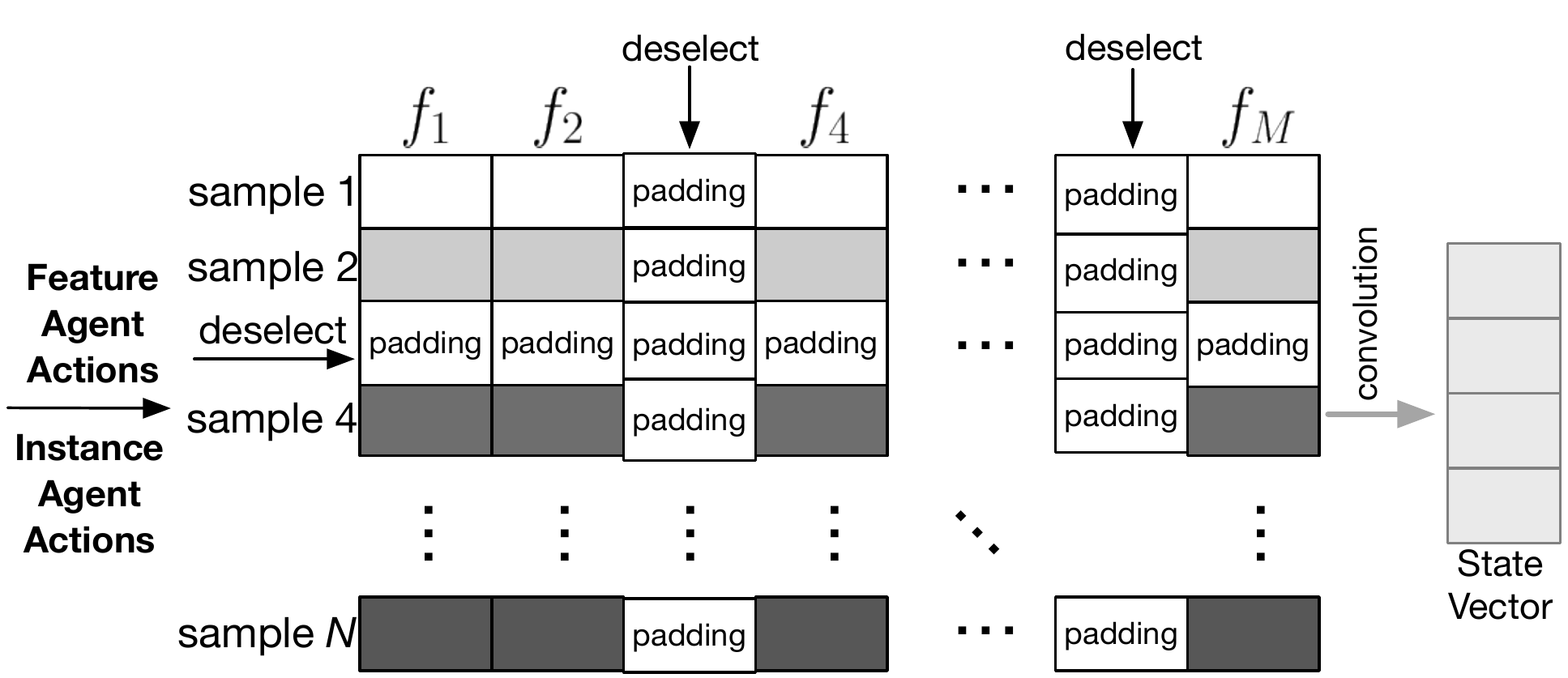}
\vspace{-3mm}
\caption{State of collaborative-changing environment, coordinating with actions of the feature agent and the instance agent.}
\label{method2}
\vspace{-1mm}
\end{figure}

To address this challenge, we develop a  dynamic state representation method inspired by the image processing technique \cite{lu2007survey}. 
Specifically, by regarding the selected data sub-matrix as a 2-dimentional image, we first fix the state dimensions by padding the deselected positions with the padding token (zero). 
Formally, for the input data $ {X}  \in {\mathbb R}^{n*m}$, at the  $t$-th step, $\mathbf{h}^t$  is measured by:
\begin{equation}
\mathbf{h}^t = 
\underbrace{ \begin{bmatrix}
{\mathbf{a_I}^t}^T || \cdots || { \mathbf{{a_I}}^t}^T  \\
\end{bmatrix}}_{m} 
\otimes
{X}
\otimes
{\underbrace{\begin{bmatrix}
{\mathbf{a_F}^t}^T || \cdots || {\mathbf{a_F}^t}^T  \\
\end{bmatrix}}_{n}}^T
\end{equation}
where $\otimes$ is element-wise product, $^T$ is the transpose of a given matrix, $\mathbf{a_I}^t$ and $\mathbf{a_F}^t$ are the action records of feature agent and instance agent at the step $t$; $m$ is the number of features; $n$ is the number of instances. 
We then utilize a single convolutional layer 
to output the final representation. Formally, the state representation at Step $t$ is computed by:
\begin{equation}
\vspace{-1mm}
\mathbf{s}^t = Conv(\mathbf{w_s} \mathbf{h}^t + \mathbf{b_s})
\end{equation}
where $\mathbf{w_s, b_s}$ is tuning weight parameters and bias, and $Conv$ is the convolution operation. Figure \ref{method2} shows the state representation process of our proposed collaborative-changing environment for dual agents.

\begin{figure*}[h]
\centering
\includegraphics[width=14cm]{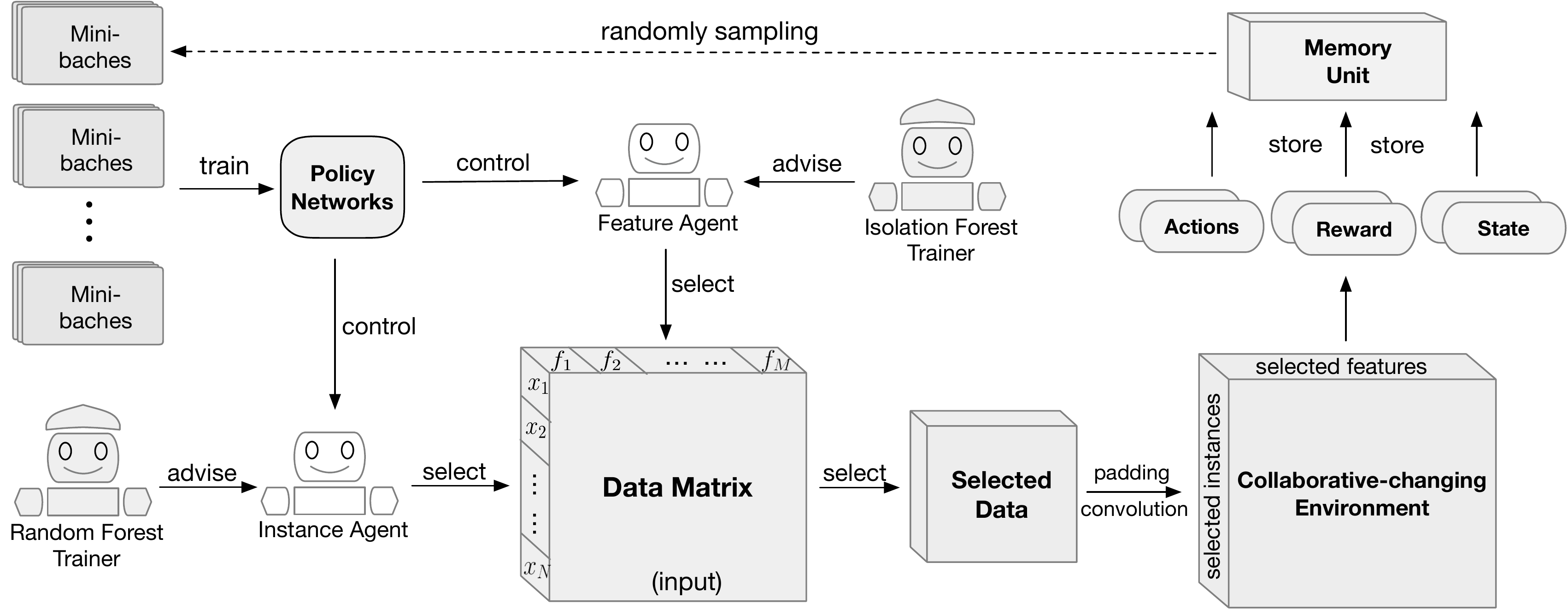}
\vspace{-1mm}
\caption{Framework Overview. Two agents collaboratively act and learn. Two trainers come to advise to help training.}
\label{overview}
\vspace{-2mm}
\end{figure*}

\subsubsection{Reward.}
The reward $r$ is to inspire the exploration of the feature agent and instance agent. 
Since the actions of dual agents are sequential scanning, we measure the reward based on the characteristic difference between last state and current state, in order to temporally train reinforcement learning decision-making process \cite{sutton2018reinforcement}. 
Specifically, supposing there are metrics set $\mathcal{K}$ we are caring about, we measure the performance difference $\Delta_{i}$ at Step $t$: 
\begin{equation}
    \Delta_{i}^t = k_i^t - k_i^{t-1}
\end{equation}
where $k_i^t$ means the $i$-th metric of the selected data subset at Step $t$.
Then, the overall reward $r$ at Step $t$ is measured by:
\begin{equation}
    r^t = \frac{1}{|\mathcal{K}|} \sum_{k_i\in\mathcal{K}}\Delta_{i}^t.
\end{equation}

The actions of feature agent and instance agent collaboratively change the state, which directly determine the reward measurement. Then the reward is shared between the two agents to inspire exploration.

\subsubsection{Trainers.}
We introduce the concept of teacher-like trainers from interactive reinforcement learning into our framework. 
We develop a random forest trainer for the feature agent and an isolation forest trainer for the instance agent. The two trainers can guide agents to explore better selection policies. More details are in the following sections.

\subsection{Model Training}
Figure \ref{overview} shows an overview of our framework. 
Dual agents have their own Deep Q-Networks (DQNs) \cite{mnih2013playing,zhang2021intelligent} as action policies. 
Two external trainers respectively guide the feature agent and the instance agent  for more efficient exploration.
The actions of the dual agents are to sequentially scan features and instances, which collaboratively decide the selected data sub-matrix $X'$. 
Then the state of the environment is derived from the sub-matrix and the reward is collected to inspire both feature agent and instance agent to select data.

To evaluate the policy, we use feed-forward neural network to approximate the Q value.
The deep Q-learning iteratively optimizes the following loss function:
\begin{equation}
     {\mathbb E} (r+\gamma \max_{a'}Q_{\theta'}(s',a')- Q_{\theta}(s,a) )^2
\end{equation}
where $r$ is the reward of action $a$ in state $s$;  
$a'$ is the next action in next state $s'$; $\gamma$ is a discount factor;
$Q_{\theta'}$ is the target network. 
{For the function approximation, the Q function is with parameter $\theta$.}
The gradient descent is:
\begin{equation}
    \theta_{t+1} \gets \theta_{t} + \alpha(r+\gamma \max_{a'}Q_{\theta'}(s',a')- Q_{\theta^t}(s,a))
\end{equation}
where $\alpha$ is the learning rate and $t$ is the training step. By optimizing $\theta$, the action policies of agents are constructed in order to maximize the long-term reward.

\subsection{Guiding Dual Agents via Interactive Reinforcement with External Trainers}
Figure \ref{overview} shows how we leverage the prior knowledge of external trainers (i.e., classical feature selection and instance filtering methods) to guide feature agent and instance agent to improve their learning efficiency.

\subsubsection{Guiding Feature Agent with Random Forest Trainer.\label{section:2}} 
Random forest classifier \cite{pal2005random} can learn a set of decision trees to measure feature importance.
We propose a trainer, namely random forest trainer, for the feature agent. The intuition is that feature importance can provide decision making support to the feature agent: if the trainer found a feature important, it would be suggested to select the feature; if not important, it would be suggested to deselect the feature.
These advice can make the feature agent more aware of characteristics of feature space. 
We develop a three-step algorithm as follows: 

\textbf{\textit{Step 1:}} We train a random forest classifier on the given $m$ features $\{\mathbf{f}_1, \mathbf{f}_2, ..., \mathbf{f}_m \}$, to obtain the importance of each feature, denoted by $ \{imp_1, imp_2, ..., imp_m\}$.

\textbf{\textit{Step 2:}} Based on the feature importance,
we design a selection probability distribution $\mathbf{p_{RF}} \{ p_1, p_2, ..., p_m \}$ for the $m$ features, where the probability for $i$-th feature is given by:
\begin{equation}
p_{i}=\left\{
\begin{array}{rcl}
1       &      & { imp_i > \frac{\beta}{m}}\\
m\,*\,imp_i       &      & {imp_i \leq \frac{\beta}{m}}
\end{array} \right. 
\end{equation}
\noindent where $\beta$ is a parameter to control suggested features.

\textbf{\textit{Step 3:}} Based on the probability, we sample an advised action list each step, denoted by $\mathbf{a_{RF}}$. The feature agent follows $\mathbf{a_{RF}}$ to take actions at the beginning steps of exploration.

\subsubsection{Guiding  Instance Agent with Isolation Forest Trainer.  \label{section:3}} 
We aim to leverage the ability of external trainers for recognizing noisy or perturbing data samples to provide advice to  instance agent for instance selection. 
Our underlying intuition is that bad instances differ from normal instances in the feature space~\cite{aggarwal2001outlier}, and an agent can follow how classic instance filtering methods identify them. 

We propose to identify bad-performed instances via an outlier detection algorithm and  let instance agent try to rule out these points.
Based on isolation forest \cite{liu2008isolation}, an outlier detector, we propose another trainer called isolation forest trainer. We show how this trainer gives advice to the instance agent step by step:


\textbf{\textit{Step 1:}} We first group instances $\{\mathbf{c}_1, \mathbf{c}_2, ..., \mathbf{c}_n \}$ based on their labels $\mathbf{Y} \{y_1, y_2, ..., y_n \}$ in classification tasks. We denote these instance groups by 
$G = \{ \mathcal{G}_{l_1}, \mathcal{G}_{l_2}, ..., \mathcal{G}_{l_p} \}$, where distinct labels $\{l_1, l_2, ..., l_p \} = set(\mathbf{Y}) $ and the group towards $k$-th distinct label $l_k$ is by $\mathcal{G}_{l_k} = \{ \mathbf{c}_i\;|\;y_i=l_k \;\&\; y_i \in \mathbf{Y}  \} $.


\textbf{\textit{Step 2:}} For each group in $G$, we utilize the isolation forest algorithm  to detect and filter outlier points. The filtered results are by $G_{IF} = \{ IF(\mathcal{G}_{l_1}), IF(\mathcal{G}_{l_2}), ..., IF(\mathcal{G}_{l_p}) \}$, where $IF$ is the filtering operation with isolation forest.

\textbf{\textit{Step 3:}} We derive the advised action list $\mathbf{A_{IF}}$ from filtered results  $G_{IF}$.
Specifically, for $ \mathbf{a_{IF}}$ $\{a_1, a_2, ..., a_n \}$ , the $i$-th advised action $a_i = 1$ if $\mathbf{c}_i \in G_{IF}$ and $a_i = 0$ if $\mathbf{c}_i \notin G_{IF} $.


\textbf{\textit{Step 4:}} At several beginning steps of the exploration in reinforcement learning, the instance agent follows the advice and take the advised actions for better training.

%% file: experiment.tex
\section{Experiment}


\begin{table*}[t]
\centering
\caption{Predictive performance of different selection methods with logistic regression as the downstream model.}
\vspace{-3mm}
\label{exp1}
\small
\begin{tabular}{c|c|c|c|c|c|c|c|c}%
\hline
\multirow{2}{*}{\diagbox{Model}{Dataset}} & \multicolumn{2}{|c}{FC} & \multicolumn{2}{|c}{Madelon} & \multicolumn{2}{|c}{Spam} & \multicolumn{2}{|c}{USPS}\\
\cline{2-9}
\multirow{3}{*}{} & {Accuracy} & {F1-score} & {Accuracy} & {F1-score} & {Accuracy} & {F1-score} & {Accuracy} & {F1-score}  \\
\hline
DROP1	&	 58.928& 59.477& 51.923& 52.048& 89.572& 89.531 & 86.559& 86.464 \\
DROP5	&	 64.682& 64.975& 52.435& 52.431& 89.717& 89.718 & 91.612& 91.607  \\
GCNN	&	 61.419& 61.460& 53.717& 53.732& 90.803& 90.846 & 93.570 &93.585\\
LASSO	&	 61.684& 62.705& 54.230& 54.223& 90.658& 90.699 & 91.899&91.937	\\
RFE  	&	 65.828& 66.430& 55.256& 55.261& 88.776& 88.862 & 93.763 &93.786 \\
LS2AOD	&	 65.057& 65.665& 53.974& 53.983& 90.948& 90.978& 93.293& 93.343 \\
GeneticFSIS& 65.834& 66.033& 55.938& 55.837&  90.742 & 90.675& 93.242&93.321 \\
IFS-CoCo &	 66.102& 65.723& 58.384& 58.323& 91.042& 91.021& 93.417& 93.519   \\
sCOs &	 66.213& 66.104& 57.992& 57.976& 90.810& 90.632& 93.792& 93.738   \\
DAIRS (Ours) &	 \textbf{67.328}& \textbf{66.908}&  \textbf{61.282} & \textbf{61.281}& \textbf{91.745}& \textbf{91.389} & \textbf{94.122} &\textbf{94.071}\\
\hline
\end{tabular}
\label{overall_table}
\vspace{-3mm}
\end{table*}
\begin{table}[h] 
\footnotesize
	\centering
	\caption{Selection ratio on different datasets.}
	\vspace{-3mm}
	\begin{tabular}{ccccc}
	\hline
		Dataset& FC&Madelon&Spam&USPS \\
	\hline
	Feature & 0.8703& 0.5980&  0.8771& 0.7187 \\
	Instance& 0.7483& 0.8829& 0.6770& 0.6266\\
	\hline
	\end{tabular}
	\label{table3}
\vspace{-4mm}
\end{table}

\subsection{Experimental Setup}

\noindent \underline{\textit{Datasets and Metrics}}: we  use  four  public  datasets of different domains  on  classification task to validate our methods: 
{\textit{ForestCover (FC)}} dataset is a publicly available dataset from Kaggle\footnote{\url{https://www.kaggle.com/c/forest-cover-type-prediction/data}} including characteristics of wilderness areas.
 {\textit{Madelon}} dataset is a Nips 2003 workshop dataset containing data points grouped in 32 clusters and labeled by 1 and -1 \cite{Dua:2019}.
{\textit{Spam}} dataset is a collection of spam emails \cite{Dua:2019}. 
 {\textit{USPS}} dataset is a handwritten digit database including handwritten digit images \cite{cai2010graph}. We report {\textit{Accuracy}} and {\noindent\textit{F1-score}} of certain downstream models to show the quality of selected data. 

\noindent \underline{\textit{Baseline Algorithms}}:
We compare predictive performance of our proposed model with different baselines, including instance selection methods, feature selection methods and feature instance joint selection methods: \textbf{DROPs.} It includes different instance reduction algorithms where we take DROP1 and DROP5 as baselines \cite{wilson2000reduction}.
\textbf{GCNN.} \cite{chou2006generalized} proposes the weak criterion employed by Condensed Nearest Neighbor.
\textbf{LASSO}. \cite{tibshirani1996regression} conducts feature selection and shrinkage via $l1$ penalty.
\textbf{RFE} (Recursive Feature Elimination) recursively deselects the least important features.
\textbf{LS2AOD} selects features via Laplacian Score and then samples via AOD.
\textbf{GeneticFSIS}. \cite{tsai2013genetic} applies genetic algorithms to alternatively select features and instances.
\textbf{IFS-CoCo}. \cite{derrac2010ifs} applies the cooperative co-evolutionary algorithm to select features and instances. 
\textbf{sCOs}. \cite{benabdeslem2020scos} uses a similarity preserving for co-selection of features and instances. For evaluation, we use these algorithms to select features/instances to get the data subset, and  evaluate the quality of selected data towards prediction. 

\noindent \underline{\textit{Implications}}:
To compare fairly, we set the downstream task as a simple classifier logistic regression. The downstream task takes selected data as input and output the classification results. We randomly split the data into train data (70\%) and test data (30\%) where categorical features are encoded in one-hot. Baseline models are set the same selection ratio as our model. 
The size of memory unit is set to 300 in experience replay. For the reward measurement, we consider accuracy, relevance score and redundancy score following \cite{liu2019automating}.
The policy networks are set as two fully-connected layers of 512 middle states with ReLU as activation function. In RL exploration, the discount factor $\gamma$ is set to 0.9, and we use $\epsilon$-greedy exploration with $\epsilon$ equal to 0.8.


\subsection{Overall Performances}

Table \ref{overall_table} shows the overall predictive performance of logistic regression on the data selected by our proposed DAIRS and compared baselines.
We report on four different datasets with respect to accuracy and f1-score and can observe our proposed model outperforms other selection methods, which signifies the best quality of our selected data for downstream predictions.
Apart from our model, we notice other joint selection methods (e.g., IFS-CoCo) can have better performance than other baselines, for these methods consider both two dimensions (feature selection and instance selection). 
\begin{figure}[h]
\vspace{-0mm}
\centering
\includegraphics[width=6.0cm]{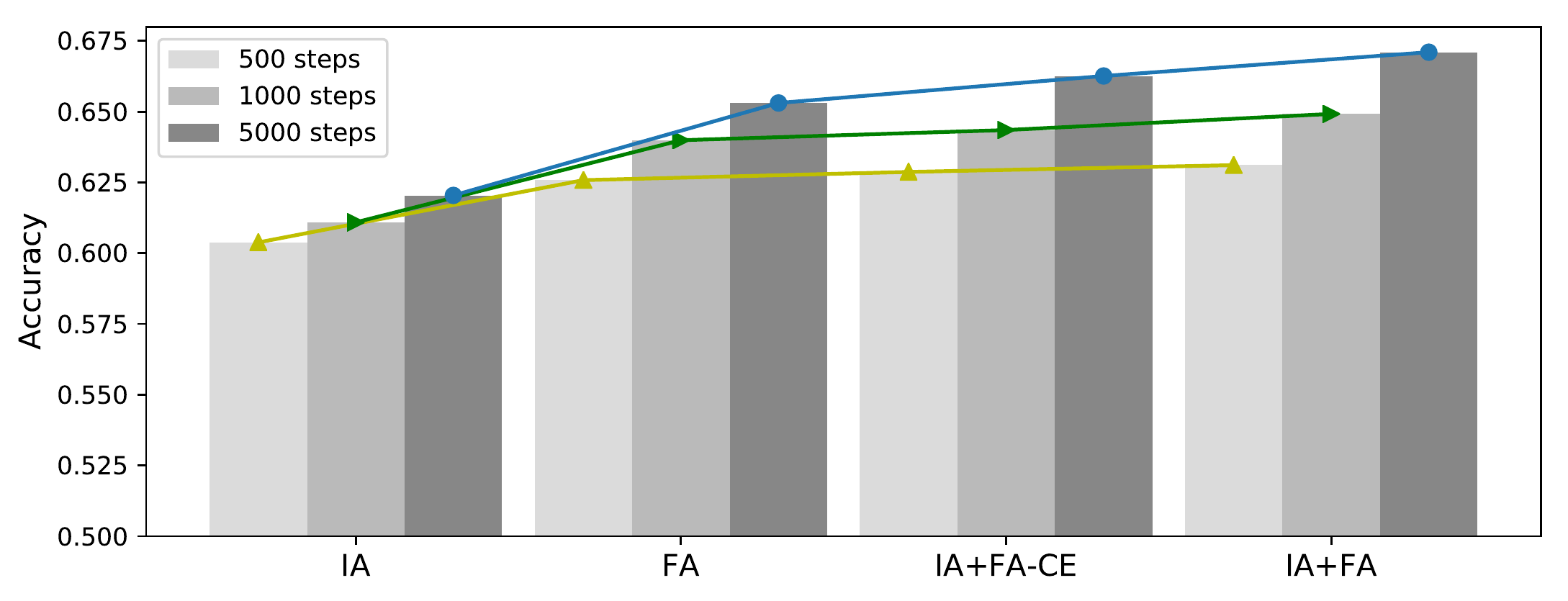}
\vspace{-4mm}
\caption{Performance of DAIRS variants on FC dataset.}
\label{exp1}
\vspace{-4mm}
\end{figure}
\begin{figure}[h]
\vspace{-0mm}
\centering
\includegraphics[width=6.0cm]{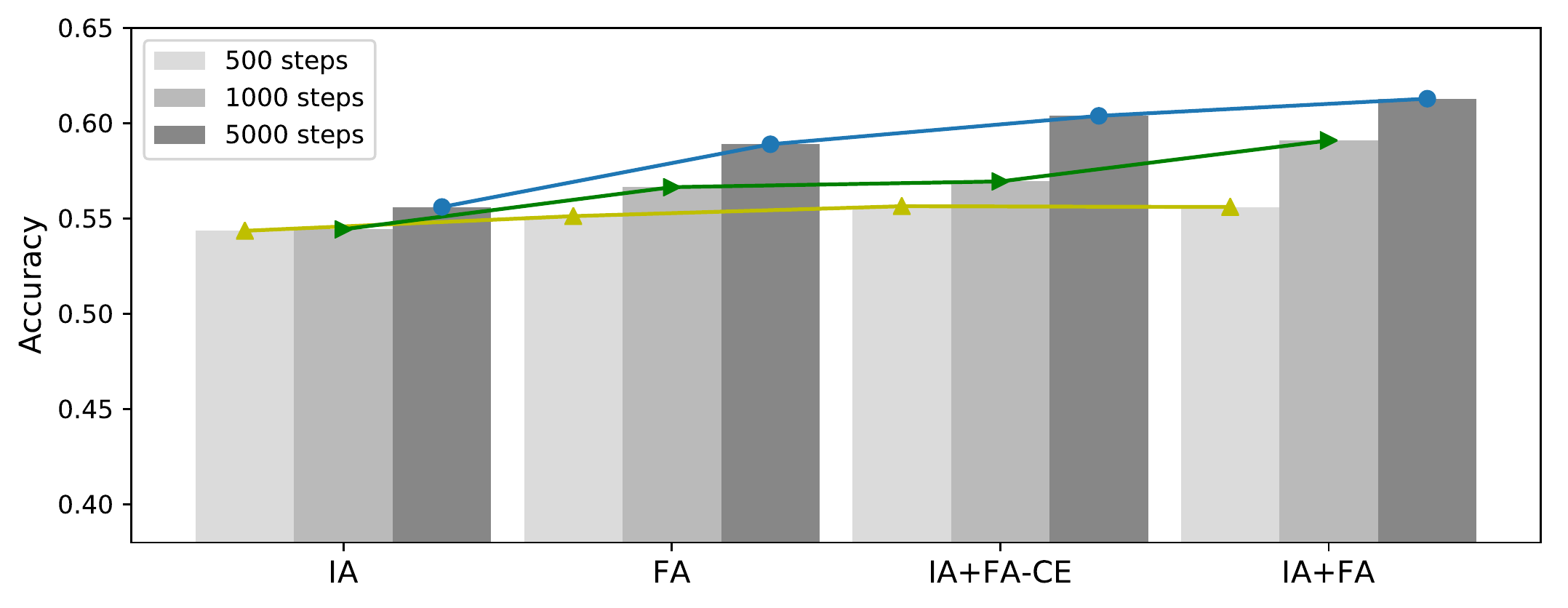}
\vspace{-4mm}
\caption{Performance of DAIRS variants on Madelon dataset.}
\label{exp2}
\vspace{-3mm}
\end{figure}
Accordingly, as it shows in Table \ref{table3}, our model can also acquire the best selection ratio to achieve higher predictive performance in exploration, while many other methods (e.g., DROP, RFE) cannot automatically learn the ratio that needs to be pre-defined. This demonstrates the superiority of our methods for automating to select the best data without handcrafted tunning.
Also, it is easy to find out in most cases, most features and instances are useful for accurate predictions, while a few ones disturb predictions.

\subsection{Study of the Dual-agent Reinforced Selection}

We aim to study the impacts of different components in our dual-agent reinforced selection framework. We consider four variant methods: (1) \textit{FA} removes the instance agent, only considers the Feature Agent and creates environment for selected features as \cite{liu2019automating}. (2) \textit{IA} removes the feature agent and only considers Instance Agent. (3) \textit{IA+FA-CE} removes Collaborative-changing Environment and creates an independent environment for each agent. (4) \textit{IA+FA} is our proposed DAIRS model with two agents. 
Figure \ref{exp1} and \ref{exp2} show performance comparison of these variants on FC and Madelon dataset. Compared to single agent's exploration, we can easily observe a performance gain by dual agents. 

Moreover, while both features and instances decide quality of the selected data, feature agent is more influential on predictions than instance agent when single agent explores. This suggests feature's quality is more important in data selection. 
We also find out the sharing environment is actually important for dual-agent coordination, which can make for better performance. From both figures, we observe the trend of blue line is more significant than yellow line; this shows our method achieves better results in long-term learning.

\subsection{Study of Interactive Reinforced Trainer}
We study the impacts of proposed trainers for interactive reinforcement learning. 
We consider four variants: (1) \textit{Non-trainer} removes both random forest trainer (RFT) and isolation forest trainer (IFT); (2) \textit{RFT} removes IFT and takes only RFT for advice; (3) \textit{IFT} removes RFT with only IFT for advice; (4) \textit{RFT+IFT} has both two trainers for advice. 

\begin{figure} \centering  
\subfigure[\small FC dataset] {
 \label{}     
\includegraphics[width=3.8cm]{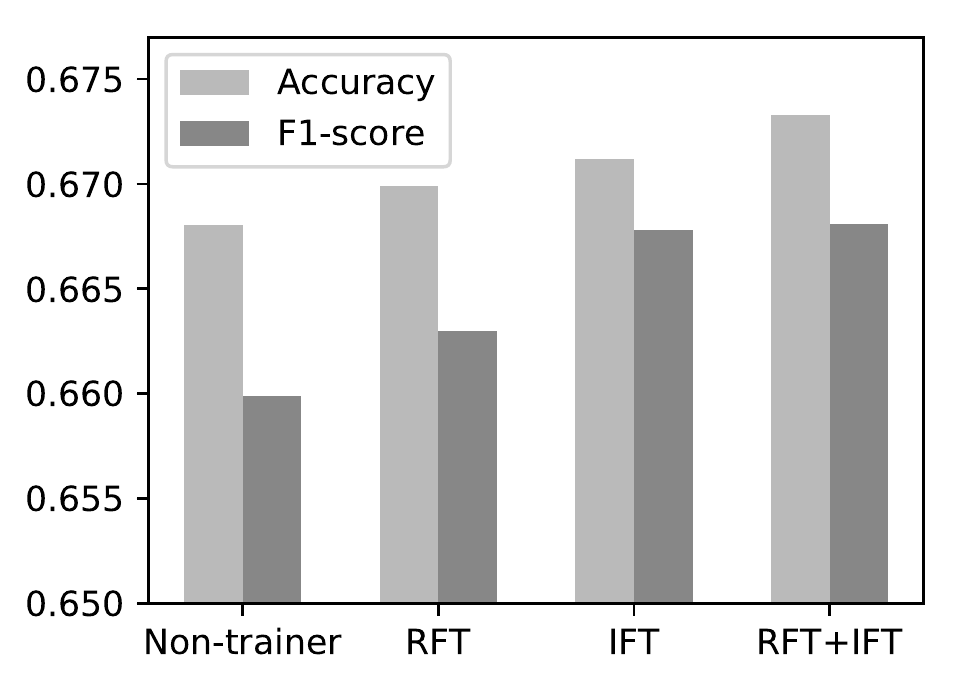} 
}  
\hspace{+1.2mm}
\subfigure[\small Madelon dataset] { 
\label{}     
\includegraphics[width=3.7cm]{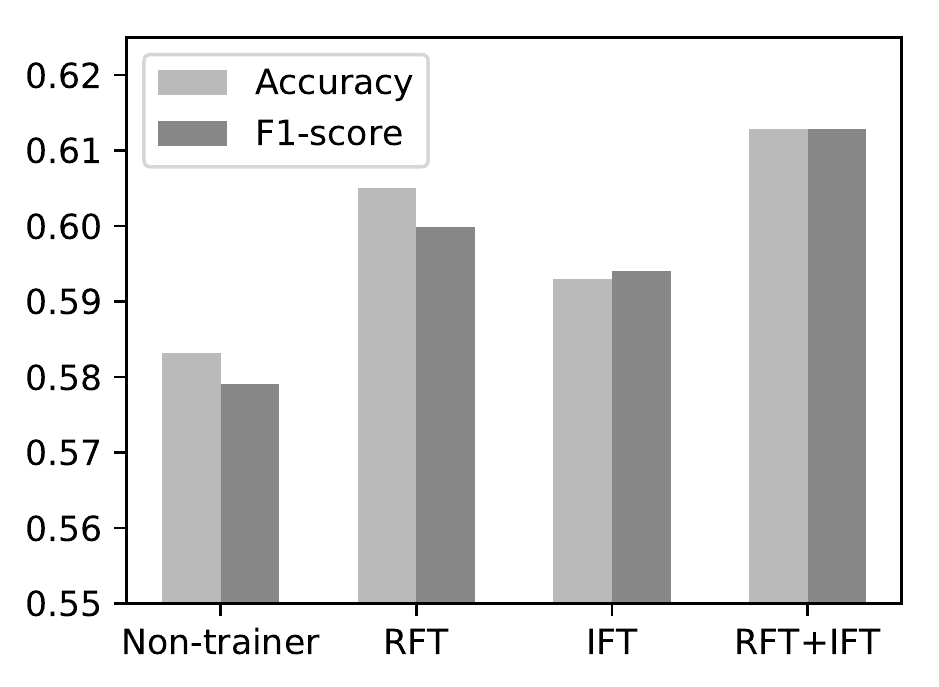}     
} 
\vspace{-3mm}
\caption{Performance of variants towards different interactive reinforced trainer setting.}     
\label{exp3}
\vspace{-6mm}
\end{figure}

\begin{figure} \centering  

\subfigure[\small FC dataset] {
 \label{}     
\includegraphics[width=3.8cm]{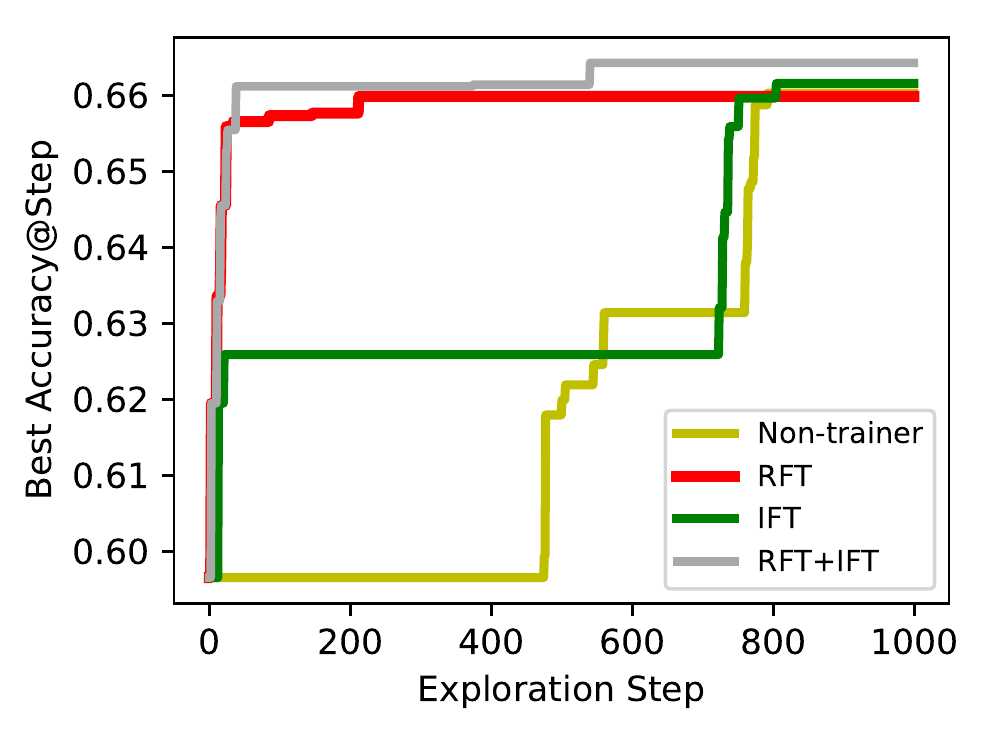} 
}  
\subfigure[\small Madelon dataset] { 
\label{}     
\includegraphics[width=3.8cm]{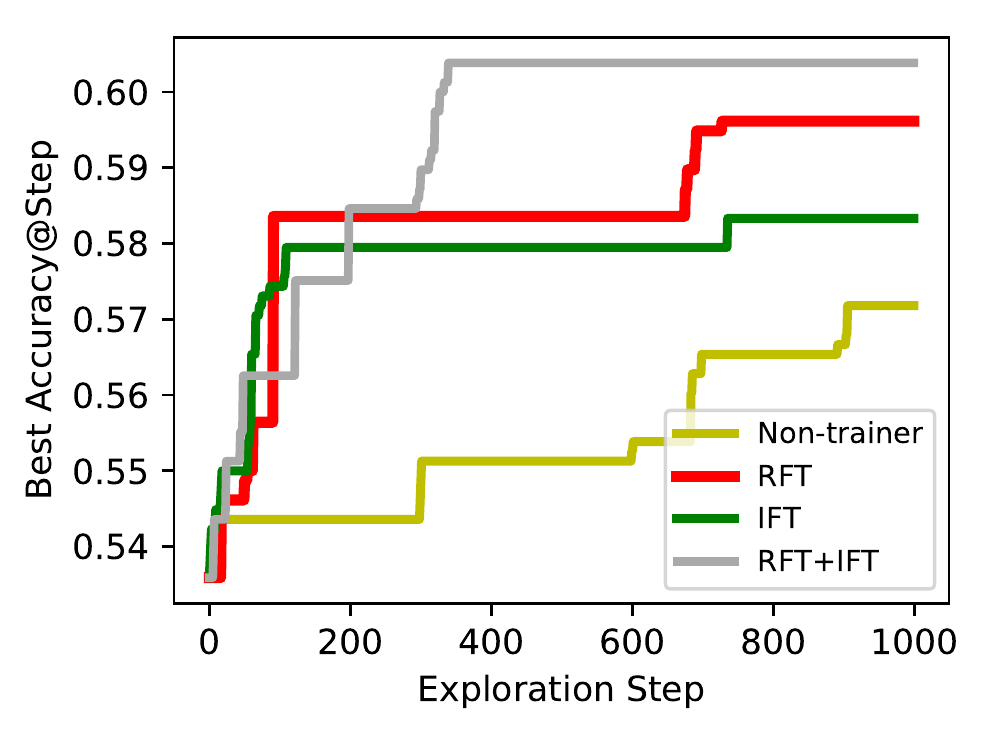}     
} 
\vspace{-3mm}
\caption{Exploration efficiency of variants with different interactive reinforced trainers.}     
\label{exp4}
\vspace{-4mm}
\end{figure}

Figure \ref{exp3} shows comparisons of variant methods. It can be observed that both two trainers help improve the data quality for better predictive performances, while the highest score is achieved when combining two trainers. 
Figure \ref{exp4} shows the efficiency comparison of each variant in terms of the explored best accuracy until current step.
The result signifies that without trainers' help, agents need more steps for exploration to achieve better results, and both trainers can help for more efficient exploration and agent learning, especially when both the feature agent and the instance agent take external advice.

\subsection{Case Study of Exploration Process}
We also try to study and visualize the exploration process of agents. Figure \ref{exp5} shows the accuracy in different steps with respect to selected feature number and instance number on FC dataset and Spam dataset. We visualize 10,000 points of exploration (colored in blue) and mark the top 10 best performed points in red.
We easily observe that most blue points are on the certain part of data space, for example the space where feature number is (30 to 50) and instance number is (5,000 to 10,000) of the left subfigure. This signifies after the initial exploration, the exploration efficiently concentrates on the optimal data subspace. 
Then, the agents can continue searching until finally finding out the best feature and instance selection results.

%% file: related_work.tex
\section{Related Work}

\textbf{Feature selection} includes three kinds of methods:
(1) Filtering methods rank features based on relevance scores and select top-ranking ones (e.g., univariate feature selection). 
(2) Wrapper methods use predictors, considering the prediction performance as objective function (e.g., branch and bound algorithms). 
(3) Embedded methods incorporate feature selection into the classifier construction to search for an optimal feature subset (e.g., LASSO \cite{tibshirani1996regression}).

\noindent \textbf{Instance selection} mainly includes two kinds of methods: (1) wrapper methods (e.g., $k$-\textit{NN} based selection); 
(2) filter methods (e.g., $kd-$trees). 
The criterion is based on the accuracy obtained by a classifier. Most of wrapper methods are based on \textit{k-NN} classifier. Other methods select instances by using SVM or Evolutionary algorithms, or are accomplished by finding border instances \cite{olvera2010review}.
\begin{figure} \centering  
\subfigure[\small FC dataset] {
 \label{}     
\includegraphics[width=4.15cm]{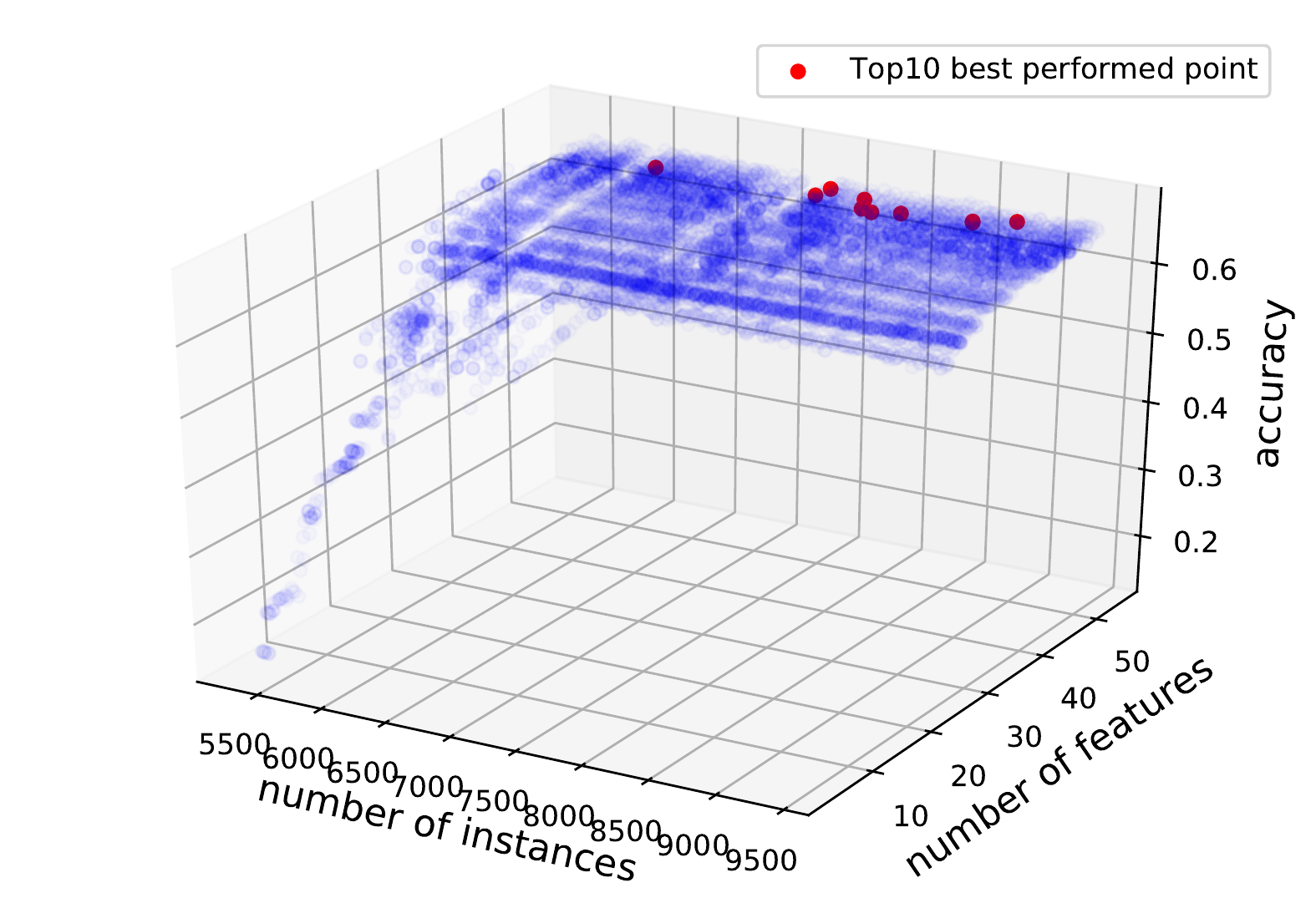} 
}  
\hspace{-4.1mm}
\subfigure[\small Spam dataset] { 
\label{}     
\includegraphics[width=4.15cm]{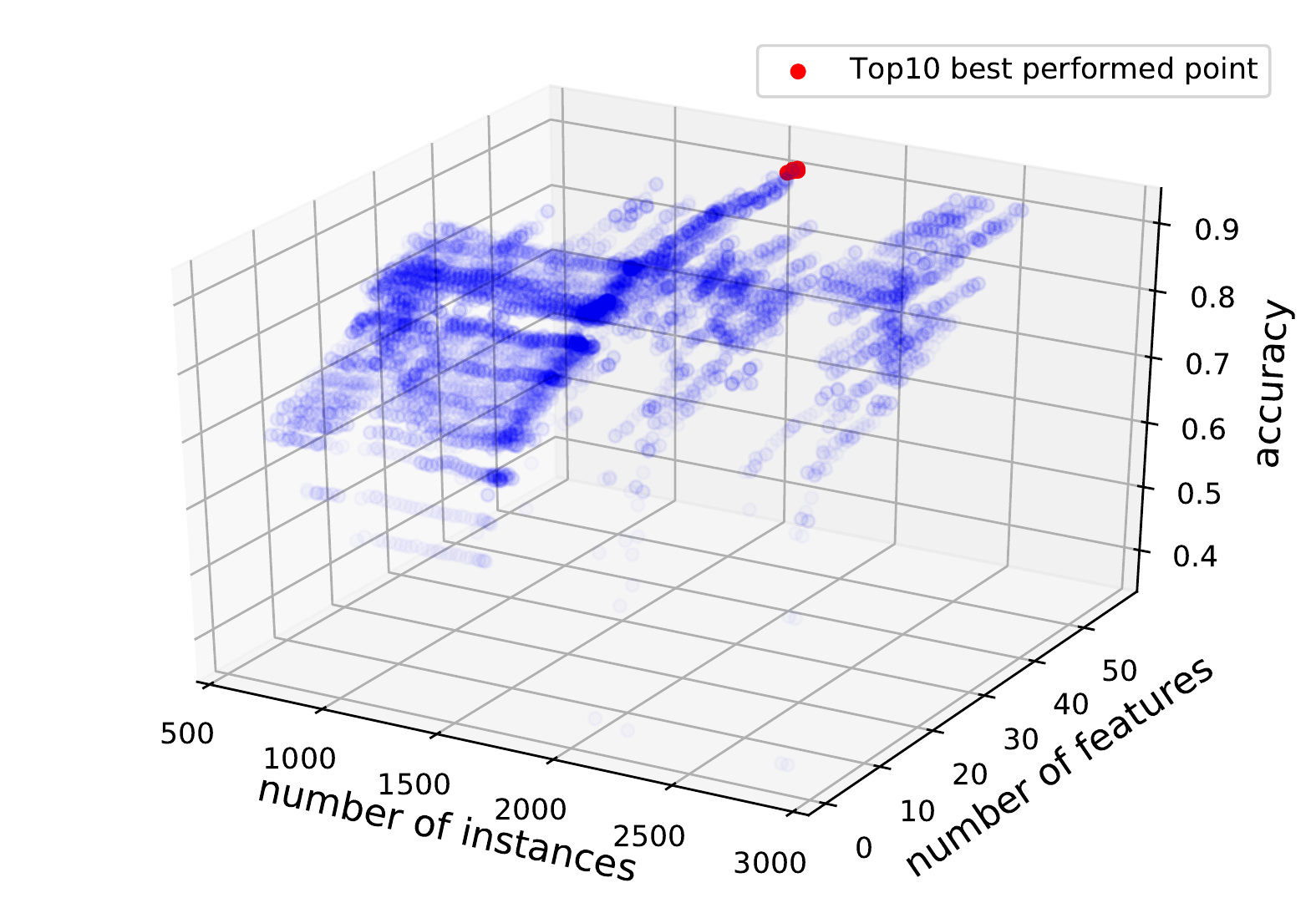}     
} 
\vspace{-2mm}
\caption{Visualizations of agent exploration process on FC and Spam dataset. Points in red are top 10 best performed steps.}     
\label{exp5}
\vspace{-4mm}
\end{figure}

\noindent \textbf{Joint selection of Feature and Instance} has also been made some efforts to study; however, limited algorithms were proposed to tackle feature and instance selection simultaneously. The joint selection is firstly studied by genetic algorithms \cite{kuncheva1999nearest}. 
Then some studies solve this problem with greedy algorithms \cite{zhang2012unified}, principal component analysis \cite{suganthi2019instance} or application of pairwise similarity \cite{benabdeslem2020scos}. Most existing work towards joint selection tries to apply heuristic search; they adopt simulated annealing algorithms \cite{de2008novel}, cooperative coevolutionary algorithms \cite{derrac2012integrating}, or sequential forward search \cite{garcia2021si}.
The joint selection is also studied in clinical data setting \cite{olvera2010review} and in social media data setting \cite{tang2013coselect}.

\noindent \textbf{Reinforced Feature Selection} has applied reinforcement learning for the feature selection task. Some studies use single agent for feature selection \cite{fard2013using,zhao2020simplifying}; others 
Multi-agent reinforcement learning have been used to automate feature selection \cite{fan2020autofs,fan2021autogfs,fan2021interactive,liu2019automating}. Inspired by these work, we  apply reinforcement learning into the joint selection task.

%% file: conclusion.tex
\vspace{-1mm}
\section{conclusion}

In this paper, we propose a dual-agent reinforcement learning framework towards the feature and instance joint selection task. We formulate the joint selection into a reinforcement learning framework with the tailor-designed sequential-scanning mechanism and collaboratively-changing environment, in order to simulate fine-grained interaction of feature space and instance space. 
The selection knowledge of the random forest trainer and the isolation forest trainer is applied to improve the efficiency of agent learning. 
The extensive experiments have demonstrated the superiority of our model on data preprocessing, which also reveals a workable design of reinforcement learning on the joint selection task.
\vspace{-1mm}
\section*{Acknowledgements}
This research was partially supported by the National Science Foundation (NSF) grant 2040950, 2006889, 2045567, IIS-2040799, IIS-2006387, IIS-1814510.

